\documentclass[conference]{IEEEtran}
\IEEEoverridecommandlockouts
\usepackage{cite}
\usepackage{amsmath,amssymb,amsfonts}
\usepackage{algorithmic}
\usepackage{graphicx}
\usepackage{textcomp}
\usepackage{xcolor}
\usepackage{amssymb}
\usepackage{tikz}
\newtheorem{theorem}{Theorem}[section]

\newtheorem{assumption}{Assumption}[section]
\newtheorem{definition}{Definition}[section]

\def\BibTeX{{\rm B\kern-.05em{\sc i\kern-.025em b}\kern-.08em
    T\kern-.1667em\lower.7ex\hbox{E}\kern-.125emX}}
\begin{document}

\title{Join-Chain Network: A Logical Reasoning View of the Multi-head Attention in Transformer\\
\thanks{ICDM 2022 Workshop on Foundation Models for Vision and Language (FOMO-VL 2022)}

}

\author{\IEEEauthorblockN{Jianyi Zhang}
\IEEEauthorblockA{
\textit{Duke University}\\
jianyi.zhang@duke.edu}
\and
\IEEEauthorblockN{Yiran Chen}
\IEEEauthorblockA{
\textit{Duke University}\\
yiran.chen@duke.edu}
\and
\IEEEauthorblockN{Jianshu Chen}
\IEEEauthorblockA{
\textit{Tencent AI Lab}\\
jianshuchen@tencent.com}}

\maketitle

\begin{abstract}
Developing neural architectures that are capable of logical reasoning has become increasingly important for a wide range of applications (e.g., natural language processing). Towards this grand objective, we propose a symbolic reasoning architecture that chains many \textit{join} operators together to model output logical expressions. In particular, we demonstrate that such an ensemble of \textit{join chains} can express a broad subset of ``tree-structured'' first-order logical expressions, named $\mathcal{FOET}$, which is particularly useful for modeling natural languages. To endow it with differentiable learning capability, we closely examine various neural operators for approximating the symbolic join-chains. Interestingly, we find that the widely used multi-head self-attention module in transformer can be understood as a special neural operator that implements the union bound of the join operator in probabilistic predicate space. Our analysis not only provides a new perspective on the mechanism of the pretrained models such as BERT for natural language understanding, but also suggests several important future improvement directions.
\end{abstract}


\section{Introduction}

Developing logical system which can naturally process symbolic rules is one of the important tasks for AI since it is a foundational model which has wide applications in language understanding and reasoning. Traditional models such as Inductive logic programming (ILP) \cite{muggleton1991inductive,muggleton1996stochastic} can learn some logical rules from a collection of positive and negative examples. However, the exponentially large searching space of the logical rules limits the scalability of tradition ILP. Considering that deep neural networks have achieved great success in many applications such as image classification, machine translation, speech recognition due to its powerful expressiveness, the question that comes naturally to us is whether we can leverage the great expressiveness power of DNNs to design the next generation logical system. Several previous attempts \cite{dong2019neural,barcelo2020logical} have been made in this direction. However, most of them are heuristic and lack clear interpretability. Developing interpretable neural architectures that are capable of logic reasoning has become increasingly important. 

Another trend in the most recent development of AI models is the wide use of multi-head attention mechanism\cite{vaswani2017attention}. Nowadays the multi-head attention has become a critical part for many foundational language and vision models, such as Bert \cite{Devlin2019BERTPO} and ViT \cite{dosovitskiy2020image}. Multiple paralleled attention heads strengthen the expressive power of a model since they can capture diverse information from different representation subspaces at different positions, which derives multiple latent features depicting the input data from different perspectives.

In our work, to develop a more interpretable neural architecture for logical reasoning, we identify a key operation for the calculation of a logic predicate. We name it as \textit{join operation}. Based on this important operation, we can convert the calculation of a logic predicate into a process of conducting the join operations recursively. We also notice that this process requires skip-connection operations pass the necessary intermediate outcomes. Based on the above observations, we design a new framework which contains several kinds of neural operators with different functions to fulfill our goal of implementing this recursive process with neural networks. We adopt the same skip-connection operation as ResNet. Interesting, for the join operator which conducts the key part of our calculation, we have found a strong connection between its operation and the mechanism of the multi-head attention. Hence, we think the widely adopted multi-head attention module can be understood as a special neural operator that implements the union bound of the join operator in probabilistic predicate space. This finding not only provides us with a good module for our logical reasoning network, but also inspires us to understand the  popular multi-head attention module in a different way, which explains its great success in language understanding. Our findings  suggest several potential directions for the improvement of the transformer \cite{vaswani2017attention}, which sheds light on the design of the large pretrained language models \cite{Devlin2019BERTPO,dosovitskiy2020image} in the future.

\begin{figure*}[ht]
    \centering
    \includegraphics[width=1\linewidth]{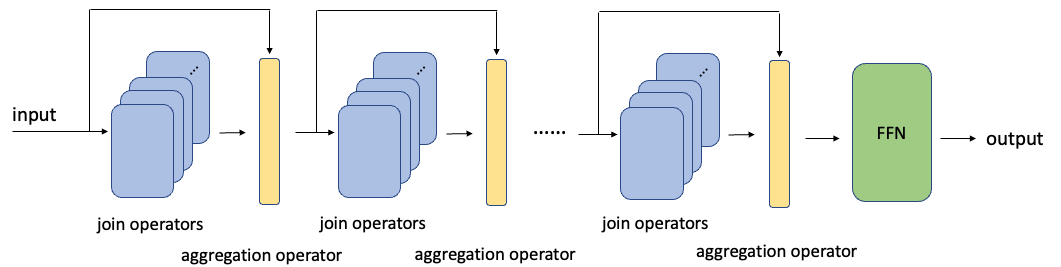}
    \caption{Skeleton of the join-chain network}
    \label{joinchainFig}
\end{figure*}

\section{Join-Chain Network}

We adopt the following example to introduce our method. 
\begin{align}\label{PNFexample}
        P(x) &= \exists y_1 \exists y_2 \exists y_{3} \exists y_{4}  \left[\overset{4} {\underset{t=1}{\wedge}} P_{t}(y_t)\wedge P_0(x)\wedge W_{(0,1)}(x,y_1)\right. \\ 
        &\left.\wedge W_{(0,2)}(x,y_2)\wedge W_{(1,3)}(y_1,y_3) \wedge W_{(1,4)}(y_1,y_4)\right]. \nonumber
\end{align}
$P(x)$ is a first-order logic predicate. To derive the value of $P(x)$ for a given $x$, we can divide the calculation into the following three steps. 

\begin{itemize}
    \item \textbf{Step 1}: We first calculate the $P_{3,1}(y_1)=\exists y_3 W_{(1,3)}(y_1,y_3) \wedge P_3(y_3)$ and $P_{4,1}(y_1)= \exists y_3 W_{(1,4)}(y_1,y_4) \wedge P_4(y_4)$.
    \item \textbf{Step 2}: We denote the $P_{new,1} (y_1)\triangleq  P_1(y_1)\wedge P_{3,1}(y_1) \wedge P_{4,1}(y_1)$. In this step, we calculate the $P_{1,0}(x) \triangleq \exists y_1 W_{(0,1)}(x,y_1) \wedge P_{new,1} (y_1)$ and $P_{2,0}(x) \triangleq \exists y_2 W_{(0,2)}(x,y_2) \wedge P_2 (y_2)$.
    \item \textbf{Step 3}: As for the final step, we need to calculate the value of $P_{1,0}(x) \wedge P_{2,0}(x)$. It is obvious that  $P(x)=P_{1,0}(x) \wedge P_{2,0}(x)$
\end{itemize}



Based on the above steps, we find a key operation for the calculation of $P(x)$ and name it as the \textbf{join operation}.
\begin{align*}
    \textit{join operation: }  P(x)= \exists yW(x, y) P(y)
\end{align*}
We can transform the calculation of $P(x)$ in Eq \ref{PNFexample} as a recursion process of the {join operations}. Hence, it is necessary to include a module in our network to calculate the join operation. Since we need to conduct the calculation of join operation in a recursive way with multiple steps, our network should have multiple layers. Besides, there are multiple join operations to conduct in each step, we need to include several join operation modules in each layer. It is worth noting that some unary predicates such as $P_2 (y_1)$ and $P_2 (y_2)$ are not used in Step 1 but thereafter used in Step 2. Hence we believe a skip connection is also necessary for our network. Based on these observations, to conduct the calculation following our steps, we design a new network which is named as the \textbf{join-chain network}. We visualize its skeleton in the Figure \ref{joinchainFig}.

As shown in the Figure \ref{joinchainFig}, the join operators will conduct the join operations on the inputs to each layer. The skip connection in each layer preserves the inputs for future use. The aggregation after the join operators will aggregate the inputs from the skip-connection and the results of join operators. This reflects the process that we need to conduct $\wedge$ operations after the join operations in each step. At the end of our framework, we adopt a feed-forward neural layer (FFN), which is designed for our last step mentioned above. 

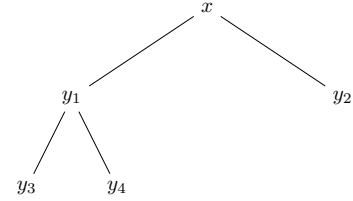
\begin{figure}[htbp!]
    \centering
    \scalebox{0.8}{
    \begin{tikzpicture} \label{ExTree}
	[thick,scale=0.5, every node/.style={scale=6}]
	\centering
	\node {$x$}
	child {node {$y_1$}
		child {node {$y_3$}
		}
		child {node {$y_4$}}
	}	
	child [missing] {}	
	child [missing] {}	
	child { node {$y_2$}
	}	
	;
	\end{tikzpicture}}
    \caption{The tree structure of example in Eq \ref{PNFexample}}
    \label{tree-examplel}
\end{figure}

For the example in Eq \ref{PNFexample} mentioned above, if we consider the set $\mathcal{N}=\{x,y_1, y_2, y_3, y_4\}$ as the node set and the index pairs set $\{(0,1),(0,2), (1,3),(1,4)\}$ as the edge set $\mathcal{E}$. We can draw the graph $\mathcal{G}=(\mathcal{N},\mathcal{E})$ in figure \ref{tree-examplel} and it is easy to find that the graph $\mathcal{G}$ is a tree. As proved in the following section, our framework can calculate the predicates which could be visualized as a graph of tree-structured. Based on the previous research, we can derive the dependency tree for every sentence. If we translate the dependency tree into a logical forms, most sentence can be represented by a logic predicate \cite{reddy2016transforming} which has a tree structure similar to the example in Figure \ref{tree-examplel}. Hence our framework has very wide applications in NLP tasks.

\section{Logical expressiveness}

Every first-order formula is logically equivalent to some formula in prenex normal form. In our study, we are interested in the case where there is only one variable $x$ which is not restricted by any quantifier. For convenience, we denote the free variable $x$ as $y_0$ and other restricted variables are denoted as $y_1, y_2,\cdots,y_T$. We assume its prenex normal form has the following formulation Eq \ref{genePNF}. 

\begin{align}\label{genePNF}
    P(x)= \exists y_1 \exists y_2 ... \exists y_T\overset{M} {\underset{m=1}{\vee}} \left[ \left(\overset{\bar{M}} {\underset{\bar{m}=1}{\wedge}} \hat{P}^m_{\bar{m}}(y_{i(m,\bar{m})})\right)\right.\\ 
    \left.\wedge \left(\overset{M^\prime} {\underset{m^\prime=1}{\wedge}} \hat{W}^m_{m^\prime}(y_{n(m,m^\prime)},y_{j(m,m^\prime)})\right)\wedge Q^m\right] \nonumber
\end{align} 
Each $\hat{P}^m_{\bar{m}}$  is a unary predicate and each $\hat{W}^m_{m^\prime}$  is a binary predicate. $Q^m$ is some {propositional constant}. $i(m,\bar{m}),n(m,m^\prime)$ and $j(m,m^\prime)$ are three index mapping functions, which map $(m,\bar{m})$ or $(m,{m}^\prime)$ to some values in $\{0, 1, 2, \cdot,T\}$. This means 
$y_{i(m,\bar{m})},y_{n(m,m^\prime)},y_{j(m,m^\prime)}\in \{y_0, y_1,...,y_T\}$. Moreover, we assume $n(m,m^\prime) < j(m,m^\prime)$. \\

\begin{theorem}\label{theorem1}
For every predicate $P(x)$ in \ref{genePNF}, there exists a $P(y_0)$ which has the following formulation and is logically equivalent to the $P(x)$.
\begin{align}\label{PNFafterfoet}
        P(y_0) = \exists y_1 \exists y_2 ... \exists y_T\overset{M} {\underset{m=1}{\vee}}  \left[\left(\overset{} {\underset{t\in {\mathcal{N}^m}}{\wedge}} P^m_{t}(y_t)\right)\right. \\ \nonumber
        \left.\wedge \left(\overset{} {\underset{(t_p,t_c)\in {{\mathcal{E}}^m}}{\wedge}} W^m_{(t_p,t_c)}(y_{t_p},y_{t_c})\right)\wedge Q^m\right],
    \end{align}
where each ${\mathcal{N}^m}  \subset \{0,1,2,\cdots, T\}$ is set of indices and $0 \in {\mathcal{N}^m} $. ${\mathcal{E}}^m$ is set of index pairs. Each ${\mathcal{N}}^m$ and ${\mathcal{E}}^m$ can form a graph $\mathcal{G}^m=({\mathcal{N}^m},{\mathcal{E}}^m)$.
\end{theorem}
Then we provide the following definitions and assumption to show the logical expressiveness of our join chain network.
\\
\begin{definition}[$\mathbf{\mathcal{FOET}(\mathcal{I}_1,\mathcal{I}_2)}$]
$\mathcal{FOET}(\mathcal{I}_1,\mathcal{I}_2)$ is the set of all the predicates $P(y_0)$ which can be transformed into the formulations in \ref{PNFafterfoet} and satisfy the following requirements.
\begin{itemize}
    \item Each $W^m_{(t_p,t_c)}(y_{t_p},y_{t_c})\in \mathcal{I}_2$ and ${P}^m_{t}(y_t)= P^m_{j_1}(y_t) \wedge P^m_{j_2}(y_t)\wedge \cdots \wedge P^m_{j_{t}}(y_t)$, where $\{ P^m_{j_1},  P^m_{j_2}, \cdots,  P^m_{j_t}\} \subset  \mathcal{I}_1$. 
    
    \item Each graph $\mathcal{G}^m=(\mathcal{N}^m,\mathcal{E}^m)$ is a tree.
\end{itemize}
\end{definition}


\begin{definition}
Each graph $\mathcal{G}^m=(\mathcal{N}^m,\mathcal{E}^m), m\in \{1,2,\cdots,M\}$ is a tree. We denote the height for each tree as $L^1, L^2, \cdots, L^{M}$ respectively and $L_{max}=\max\{L^1, L^2, \cdots, L^{M}\}$.  We denote the number of leaf nodes of each tree as $H^1, H^2, \cdots, H^{M}$ respectively and $H_{sum}=H^1+ H^2+ \cdots+ H^{M}$. We define the \textbf{height} of the predicate as $L=L_{max}$ and the \textbf{width} of the predicate as $H=H_{sum}$. 
\end{definition}

\begin{definition}[${\mathcal{FOET}_{\{\bar{L},\bar{H}\}}(\mathcal{I}_1,\mathcal{I}_2)}$]$\mathcal{FOET}_{\{\bar{L},\bar{H}\}}(\mathcal{I}_1,\mathcal{I}_2)$ is the set of the predicates $P(x)\in \mathcal{FOET}$ with height $L\leq \bar{L}$ and width $H\leq \bar{H}$.
\end{definition}

\begin{assumption}\label{assumption}
For any $W \in \mathcal{I}_2$, there exists some function $f$, such that $W=f(I_s)$, where $I_s \subset \mathcal{I}_1$.\\
\end{assumption}
Now we can provide the theorem to show the logical expressiveness.\\
\begin{theorem}\label{theorem2}
Under the assumption \ref{assumption}, a join-chain network with the $\bar{H}$-head and $\bar{L}$-layer self-attention block can express all the predicates in ${\mathcal{FOET}_{\{\bar{L},\bar{H}\}}(\mathcal{I}_1,\mathcal{I}_2)}$ if the input to the join-chain network is $\mathcal{I}_1$. \\
\end{theorem} 

We look back at the example in Eq \ref{PNFexample} to understand the definitions and theorem. There is one tree \textit{i.e.} M=1. The node set of the graph is $\mathcal{N}=\{0,1,2,4\}$. The edge set of the graph $\mathcal{E}$ is $\{(0,1),(0,2),(1,3),(1,4)\}$. The graph $(\mathcal{N},\mathcal{E})$ is a tree visualized in the Figure \ref{tree-examplel}. The height of the $P(y_0)$ is 2. The width is 3 since there are 3 leaf nodes \textit{i.e.} $y_3$, $y_4$ and $y_2$. According to the theorem \ref{theorem2}, $P(x)$ can be expressed by the join-chain network with 3 heads and 2 layers. 


\section{rethink multi-head attention as join operation}

To design the join operator with differentiable learning capability, we study various neural operators for approximating the symbolic join operations. Interestingly, we find that the widely adopted multi-head attention mechanism can be understood as a join operator. 

First, we denote the domain of all the predicates, including all the binary predicates $W(x,y)$ and unary predicates $P(y)$, as $\{x_1, x_1, x_3, ..., x_S\}$, which means $x, y \in \{x_1, x_2, x_3, ..., x_S\}$. Then in the multi-head attention mechanism, the core part is the product between the self-attention matrix $\mathbf{A}$ and the value tensor $\mathbf{V}$, \begin{align}\label{self-attention}
\mathbf{Z}=\mathbf{A} \mathbf{V}
\end{align}

If we want to calculate the the join operation between $W(x,y)$ and $P(y)$, we know
\begin{align}
    \exists y W(x,y)\wedge P(y)=\overset{S} {\underset{s=1}{\vee}} W(x,x_s)\wedge P(x_s)
\end{align}
Then, if the multiplication can be understood as the conjunction operation $\wedge$ and the addition as the disjunction $\vee$, we can consider the value tensor V as $[P(x_1), P(x_2),..., P(x_S)]$ and self-attention matrix $A$ as $\{W(x_s, x_{s\prime})\}$. Based on this,  the $s$-th element $z_s$ in the tensor $Z$ is the value of $\exists y W(x_s,y)\wedge P(y)$. Hence the tensor $Z$ will the join operation between $W(x,y)$ and $P(y)$. Generally speaking, the self-attention matrix $A$ learns all the values of the binary predicate $W(x,y)$, and the value tensor $V$ learns all the values of the unary predicate $P(y)$. For each head of attention mechanism in each layer, the self-attention matrix $\mathbf{A}$ learns a binary predicate $W(x,y)$. Hence, the amount of the leaf nodes in the Figure \ref{tree-examplel} reflects the importance of multiple heads for self-attention.

\section{Discussion}
Our work provides a novel understanding of the multi-head attention from the logical reasoning view. Based on the previous work on semantic parsing such as dependency tree and lambda dependency-based compositional semantics \cite{reddy2016transforming,liang2013lambda}, most sentences can be represented in logical forms which have similar tree structure as the example in Eq \ref{PNFexample}. Hence, our work provides a novel explanation why the multi-head attention achieves great success in recent development of NLP from a new perspective. Furthermore, the logic reasoning view also provides us with some suggestions on how to improve the design of transformers. Since logical expressions of most sentences in NLP have tree structures similar to the example in Eq \ref{PNFexample} shown in Figure \ref{tree-examplel}, the number of join operations decreases as we proceed the calculation. This means the amount of the heads could decrease as the layer become less close to the inputs. This provides us with a new insight on how to compress the multi-head attention blocks in transformers. Besides, it is worth noting that the skip connection is also heavily utilized in the transformers. Our work provide a new interpretation for the use of skip-connection in transformers which is different from its original motivation in residual learning. Moreover, the assumption \ref{assumption} also provides us with a potential way to augment the transformer with some external knowledge. We could incorporate some additional commonsense knowledge into the self-attention block to boost the logical reasoning as well as the inference of transformer. 

Our work has lots of interesting future directions. One is to design some more efficient neural operators for the join operations, which could be valuable for both improving logical reasoning and multi-head attention in transformers. Another important direction is to enhance the expressiveness capacity of our model. We hope it could handle the logical predicates which do not have the tree structures. Besides, how to enable our model to process the logical predicates which depicts the relationship between three or more variables is also very challenging but interesting.

\section{Conclusion}

We identify a key operation \textit{i.e.} join operation for logical reasoning, which guides us to propose a new interpretable neural network for this task. We also prove its logical expressiveness. Interestingly, we find the multi-head attention mudules can be understood as a neural operator for  join operation. Our work provides a new understanding of the multi-head attention mechanism in transformers and sheds light on how to improve the recent pretrained models which adopt the multi-head attention mudules.

\onecolumn
\newpage

\appendix

\subsection{Proof of Theorem \ref{theorem1} }

Every first-order formula is logically equivalent to some formula in prenex normal form. In our study, we are interested in the case where there is only one variable $x$ which is not restricted by any quantifier. For convenience, we denote the free variable $x$ as $y_0$ and other restricted variables are denoted as $y_1, y_2,\cdots,y_T$. We assume its prenex normal form has the following formulation. 

\begin{align}\label{genePNFapp}
    P(x) = \exists y_1, \exists y_2, ..., \exists y_T\overset{M} {\underset{m=1}{\vee}} \left[ \left(\overset{\bar{M}} {\underset{\bar{m}=1}{\wedge}} \hat{P}^m_{\bar{m}}(y_{i(m,\bar{m})})\right)\wedge \left(\overset{M^\prime} {\underset{m^\prime=1}{\wedge}} \hat{W}^m_{m^\prime}(y_{n(m,m^\prime)},y_{j(m,m^\prime)})\right)\wedge Q^m\right]
\end{align} 
Each $\hat{P}^m_{\bar{m}}$ is some unary predicate and each $\hat{W}^m_{m^\prime}$ is some binary predicate \textbf{but not tautology}.$Q^m$ is some propositional constant. $i(m,\bar{m}),n(m,m^\prime)$ and $j(m,m^\prime)$are three index mapping functions, which map $(m,\bar{m})$ or $(m,{m}^\prime)$ to some value in $\{0, 1, 2, \cdot,T\}$. This means 
$y_{i(m,\bar{m})},y_{n(m,m^\prime)},y_{j(m,m^\prime)}\in \{y_0, y_1,...,y_T\}$. Moreover, we assume $n(m,m^\prime) < j(m,m^\prime)$.

\subsubsection{Transformation}

\begin{definition}[\textbf{Transformation A}]
Transformation A is defined as the transformation from \ref{genePNFapp} to \ref{PNFafterapp}.
\end{definition}

\begin{itemize}
    \item[Step 1] For each $m$, if there exists $\{\bar{m}_1, \bar{m}_2, \cdots, \bar{m}_q\}$ such that 
    $i(m,\bar{m}_1)=i(m,\bar{m}_2)=\cdots=i(m,\bar{m}_q)=t$, we conduct the conjunction on the corresponding $\hat{P}^m_{\bar{m}_1},\hat{P}^m_{\bar{m}_2},\cdots, \hat{P}^m_{\bar{m}_q}$ and derive the following new predicate $P^m_{t}$
    \begin{align}\label{step1}
       P^m_{t}(y_t)=\hat{P}^m_{\bar{m}_1}(y_t)\wedge \hat{P}^m_{\bar{m}_2}(y_t)\wedge\cdots\wedge \hat{P}^m_{\bar{m}_q}(y_t)
    \end{align}
    Then the \ref{genePNFapp} can be transformed into the following formulation
    \begin{align}\label{step1}
        P(x) = \exists y_1, \exists y_2, ..., \exists y_T\overset{M} {\underset{m=1}{\vee}}  \left[\left(\overset{} {\underset{t\in \widetilde{\mathcal{N}}_m}{\wedge}} P^m_{t}(y_t)\right)\wedge \left(\overset{M\prime} {\underset{m\prime=1}{\wedge}} \hat{W}^m_{m^\prime}(y_{n(m,m^\prime)},y_{j(m,m^\prime)})\right)\wedge Q^m\right],
    \end{align}
     where $\widetilde{\mathcal{N}}^m$ is the set of indices and $\widetilde{\mathcal{N}}^m \subset \{1,2,\cdots, T\}$
    \item[Step 2] For each $m$, if there exists $(m^\prime_1, m^\prime_2, \cdots, m^\prime_q)$ such that $n(m,m^\prime_1)=n(m,m^\prime_2)=\cdots=n(m,m^\prime_q)=t_c$ and  $j(m,m^\prime_1)=j(m,m^\prime_2)=\cdots=j(m,m^\prime_q)=t_p$, we conduct the conjunction on the corresponding $\hat{W}^m_{m^\prime_1}, \hat{W}^m_{m^\prime_2},\cdots, \hat{W}^m_{m^\prime_q}$ and derive the following new predicate $W^m_{(t_p,t_c)}$
    \begin{align}
        W^m_{(t_p,t_c)}(y_{t_p},y_{t_c})=\hat{W}^m_{m^\prime_1}(y_{t_p},y_{t_c})\wedge \hat{W}^m_{m^\prime_2}(y_{t_p},y_{t_c})\wedge \cdots \wedge \hat{W}^m_{m^\prime_q}(y_{t_p},y_{t_c})
    \end{align}
    After this transformation, \ref{step1} has the following formulation
    \begin{align}\label{PNFmiddleapp}
        P(x) = \exists y_1, \exists y_2, ..., \exists y_T\overset{M} {\underset{m=1}{\vee}}  \left[\left(\overset{} {\underset{t\in \widetilde{\mathcal{N}}^m}{\wedge}} P^m_{t}(y_t)\right)\wedge \left(\overset{} {\underset{(t_p,t_x)\in \hat{\mathcal{E}}^m}{\wedge}} W^m_{(t_p,t_c)}(y_{t_p},y_{t_c})\right)\wedge Q^m\right],
    \end{align}
 where $\hat{\mathcal{E}}^m$ is the set of index pairs $(t_p,t_c)$.
    
        \item[Step 3] We will move to a further discussion on the set $\hat{\mathcal{E}}^m$. We can consider $(t_p,t_c)$ as an edge which connects node $t_p$ and node $t_c$, we denote the set of all the nodes which appear in the edge set $\hat{\mathcal{E}}^m$ is $\hat{\mathcal{N}}^m$. 
    
 If the index "0", which corresponds to the free variable $y_0$, is not included in $\hat{\mathcal{N}}^m$, we need to add an edge $(0,t_{min})$ to the $\hat{\mathcal{E}}^m$, where $t_{min}$ is the smallest index in $\hat{\mathcal{N}}^m$. 
 
 For all the nodes $\{t_1, t_2, t_3, \cdots, t_s\}$ which are in $\widetilde{\mathcal{N}}^m$ but not in $\hat{\mathcal{N}}^m\cup \{0\}$, we need to add the edges $\{(0,t_1), (0,t_2), (0,t_3), \cdots, (0,t_s)\}$
 to $\hat{\mathcal{E}}^m$. We can derive the following edge set and node set
 \begin{align*}
     {\widetilde{\mathcal{E}}^m}&=\hat{\mathcal{E}}^m\cup \{(0,t_{min})\}\cup \{(0,t_1), (0,t_2), (0,t_3), \cdots, (0,t_s)\}\\
     {\mathcal{N}^m}&=\hat{\mathcal{N}}^m \cup\{0\}\cup \{t_1, t_2, t_3, \cdots, t_s\}
 \end{align*}
 
 After this, we introduce the third step of \textbf{Transformation A}. We can derive the following formulation which is logical equivalent to \ref{PNFmiddleapp}.
  \begin{align}\label{PNFafterapp}
        P(x) = \exists y_1, \exists y_2, ..., \exists y_T\overset{M} {\underset{m=1}{\vee}}\left[\left(\overset{} {\underset{t\in {\mathcal{N}^m}}{\wedge}} P^m_{t}(y_t)\right)\wedge \left(\overset{} {\underset{(t_p,t_c)\in \widetilde{\mathcal{E}}^m}{\wedge}} W^m_{(t_p,t_c)}(y_{t_p},y_{t_c})\right)\wedge Q^m \right]
    \end{align}
    where $W^m_{(t_p,t_c)}\equiv 1$ for all the $(t_p,t_c)\in {\widetilde{\mathcal{E}}^m} \backslash \hat{\mathcal{E}}^m$ and $P^m_{t}\equiv 1 $ for all $t \in {\mathcal{N}^m} \backslash \hat{\mathcal{N}}^m$

\end{itemize}

\textbf{Transformation A} : Transformation from \ref{genePNFapp} to \ref{PNFafterapp} preserves the \textbf{Logically Equivalence}.

With these node set ${\mathcal{N}}^m$ and edge set ${\widetilde{\mathcal{E}}}^m$, we can derive a graph $\widetilde{G}^m=({\mathcal{N}}^m,{\widetilde{\mathcal{E}}}^m)$ for each $m$. If each graph ${G}^m$ is composed of several trees, we say the predicate is \textbf{Tree-structured}.
 
\begin{definition}[\textbf{Transformation B}]
Transformation B is defined as the following transformation from \ref{PNFafterapp} to \ref{PNFafterfoetapp}.
\end{definition}

For each $m$, the graph derived with ${\mathcal{N}^m}$ and ${{\mathcal{E}}^m}$ is composed of several trees. Since $\{0\}\subset \mathcal{N}^m$, $0$ is always one of the root nodes. We denote the the other root nodes of the these trees as $\{{r_1},{r_2},\cdots,{r_{v_m}}\}\subset \mathcal{N}^m \subset \{1,2,\cdots,N\}$. We need to add the edges $\{(0,r_1), (0,r_2), (0,r_3), \cdots, (0,r_{v_m})\}$ to ${\mathcal{E}}^m$ and derive the new edge set for each $m$.
\begin{align*}
    {{\mathcal{E}}^m}=\widetilde{\mathcal{E}}^m \cup \{(0,r_1), (0,r_2),\cdots, (0,r_{v_m})\}
\end{align*}
Then we can derive the following formulation.
\begin{align}\label{PNFafterfoetapp}
        P(y_0) = \exists y_1, \exists y_2, ..., \exists y_T\overset{M} {\underset{m=1}{\vee}}  \left[\left(\overset{} {\underset{t\in {\mathcal{N}^m}}{\wedge}} P^m_{t}(y_t)\right)\wedge \left(\overset{} {\underset{(t_p,t_c)\in {{\mathcal{E}}^m}}{\wedge}} W^m_{(t_p,t_c)}(y_{t_p},y_{t_c})\right)\wedge Q^m\right]
    \end{align}
    where $W^m_{(t_p,t_c)}\equiv 1$ for all the $(t_p,t_c)\in {\mathcal{E}^m} \backslash \widetilde{\mathcal{E}}^m$.

Thereafter, the trees mentioned above can be combined together as one tree. That means the graph $G^m(\mathcal{N}^m,\mathcal{N}^m)$ will be a tree.

\subsection{{Proof of Theorem \ref{theorem2} }}

\textbf{The Simple Case}
We start from a simple case where 
\begin{align}
    \mathcal{E}^1=\mathcal{E}^2=\cdots=\mathcal{E}^{M} \triangleq \mathcal{E}.
\end{align}

And for each $(t_p,t_c)\in {\mathcal{E}^m}$ 
\begin{align*}
    W^1_{(t_p,t_c)}\equiv  W^2_{(t_p,t_c)} \equiv \cdots \equiv  W^{M}_{(t_p,t_c)}\triangleq W_{(t_p,t_c)}
\end{align*}

In this case, all the trees share the same structure and the corresponding $W^i_{(t_p,t_c)}$ in each tree are the same. Then it is easy to verify that 
\begin{align}
    \mathcal{N}^1=\mathcal{N}^2=\cdots=\mathcal{N}^{M} \triangleq \mathcal{N}.
\end{align}

We first derive a partition on $\mathcal{N}$ through the following operation.

Assuming there are $H$ leaf nodes, the first group is the set of all the leaf nodes. We denote this set as $\mathcal{N}_1$ and each leaf node as $t(1,h)$ where $1 \leq  h \leq H$. 
\begin{align*}
    \mathcal{N}_1=\{t(1,h)|1\leq h\leq H\}\subset \{1,2,\cdots,T\}
\end{align*}

Furthermore, we also denote the parent node which connects to the leaf node as $s(1,h)$ where $1 \leq  h \leq H$ and the set of these parent nodes as $\mathcal{S}_1$. \textbf{Here we abuse the definition of set since there might be duplicate elements in $\mathcal{S}_1$.}
\begin{align}
    \mathcal{S}_1=\{s(1,h)|1\leq h\leq H\}\subset \{1,2,\cdots,T\}\backslash \mathcal{N}_1
\end{align}   
The set of all the edges which connect the leaf nodes and their own father nodes is denoted as $\mathcal{E}_1$.
\begin{align}
    \mathcal{E}_1=\{(s(1,h),t(1,h)|1\leq h\leq H\}\subset \mathcal{E}.
\end{align}

Then we eliminate all the leaf nodes $t(1,h)$ in  $\mathcal{N}_1$. We assume there are $H_2$ leaf nodes thereafter. The second group $\mathcal{N}_2$ is the set of all the leaf nodes $t(2,h)$ where $1 \leq  h \leq H_2$. The second group $\mathcal{S}_2$ is the set of all the parent nodes $s(2,h)$ which connects to the leaf node $t(2,h)$. The second group $\mathcal{E}_2$ is set of all the edges which connect them.
\begin{align}
    &\mathcal{N}_2=\{t(2,h)|1\leq h\leq H_2\}\subset \{1,2,\cdots,T\}\backslash \mathcal{N}_1\\
    &\mathcal{S}_2=\{s(2,h)|1\leq h\leq H_2\}\subset \{1,2,\cdots,T\}\backslash (\mathcal{N}_1\cup\mathcal{N}_2)\\
   & \mathcal{E}_2=\{(s(2,h),t(2,h)|1\leq h\leq H_2\}\subset \mathcal{E}\backslash\mathcal{E}_1
\end{align}

Iteratively, after we eliminate all the leaf nodes $t(l-1,j)$ in the $(l-1)$-th set $\mathcal{N}_{l-1}$, we assume there are $h_{l}$ leaf nodes thereafter. The $l$-th group $\mathcal{N}_l$ is the set of all the leaf nodes $t(l,j)$ where $1 \leq  j \leq h_1$. The $l$-th group $\mathcal{S}_l$ is the set of all the parent nodes $s(l,j)$ which connects to the leaf node $t(l,j)$. The $l$-th group $\mathcal{E}_l$ is set of all the edges which connect them.
\begin{align}
    &\mathcal{N}_l=\{t(l,h)|1\leq h\leq H_l\}\subset \{1,2,\cdots,T\}\backslash (\mathcal{N}_1\cup\mathcal{N}_2\cup \cdots \cup\mathcal{N}_{l-1}) \\
    &\mathcal{S}_l=\{s(l,h)|1\leq h\leq H_2\}\subset \{1,2,\cdots,T\}\backslash (\mathcal{N}_1\cup\mathcal{N}_2\cup \cdots \cup\mathcal{N}_l )\\
   & \mathcal{E}_l=\{(s(l,h),t(l,h))|1\leq h\leq H_2\}\subset \mathcal{E}\backslash(\mathcal{E}_1\cup\mathcal{E}_2 \cdots \cup\mathcal{E}_{l-1})
\end{align}

Since the depth of the tree is $L$, there are $L$ groups $\{\mathcal{N}_1,\mathcal{N}_2,\cdots,\mathcal{N}_L\}$. According to the definition of each $N_l$, if $l_1 \neq l_2$, we have 
\begin{align*}
    &\mathcal{N}=\mathcal{N}_1 \cup \mathcal{N}_2 \cup \cdots \cup \mathcal{N}_L \\
    &\mathcal{N}_{l_1} \cap \mathcal{N}_{l_2} = \emptyset
\end{align*}

Hence, $\{\mathcal{N}_1,\mathcal{N}_2,\cdots,\mathcal{N}_L\}$ is a partition of $\mathcal{N}$. Similarly, $\{\mathcal{E}_1,\mathcal{E}_2,\cdots,\mathcal{E}_{L-1}\}$ is a partition of $\mathcal{E}$.

\begin{align}\label{tragetFOET}
    P(y_0) = \exists y_1, \exists y_2, ..., \exists y_7\overset{2} {\underset{m=1}{\vee}}  \left[\left(\overset{} {\underset{t\in {\mathcal{N}^m}}{\wedge}} P^m_{t}(y_t)\right)\wedge \left(\overset{} {\underset{(t_p,t_c)\in {{\mathcal{E}}^m}}{\wedge}} W^m_{(t_p,t_c)}(y_{t_p},y_{t_c})\right)\wedge Q^m\right]
\end{align}
,where
\begin{align*}
     &\mathcal{N}=\mathcal{N}^1=\mathcal{N}^2=\{0,1,2,3,4,5,6,7\}\\
    &\mathcal{E}=\mathcal{E}^1=\mathcal{E}^2=\{(0,1),(0,2),(0,3),(1,4),(1,5),(2,6),(5,7)\}
\end{align*}
According to the discussion above we have the following groups.

\begin{align*}
    &\mathcal{N}_1=\{4,7,6,3\}, \mathcal{S}_1=\{1,5,2,0\},
     \mathcal{E}_1=\{(1,4),(5,7),(2,6),(0,3)\}\\
         &\mathcal{N}_2=\{5,2\},
     \mathcal{S}_2=\{1,0\},\mathcal{E}_2=\{(1,5),(0,2)\}\\
     &\mathcal{N}_3=\{1\},\mathcal{S}_2=\{0\},\mathcal{E}_2=\{(0,1)\}\\
     & \mathcal{N}_4=\{0\}
\end{align*}

Our target is 
\begin{align}\label{PNFtarget}
        P(y_0) = \exists y_1, \exists y_2, ..., \exists y_T \overset{M} {\underset{m=1}{\vee}}  \left[\left(\overset{} {\underset{t\in {\mathcal{N}}}{\wedge}} P^m_{t}(y_t)\right)\wedge \left(\overset{} {\underset{(t_p,t_c)\in {\mathcal{E}}}{\wedge}} W_{(t_p,t_c)}(y_{t_p},y_{t_c})\right)\wedge Q^m\right]
\end{align}

\textbf{The First Layer}

Similar to the proof for toy case, the first layer eliminate all the leaf nodes in $\mathcal{N}_1$. There are $H$ heads. 

For the $h$-th head in the first layer, the value matrix will represent the set of predicates $\mathcal{V}_{t(1,h)}=\{P^{1}_{t(1,h)},P^{2}_{t(1,h)},\cdots,P^{M}_{t(1,h)}\}$. 
Then the $k$-th head will learn the join operation between $W_{(s(1,h),t(1,h))}$ and $\mathcal{V}_{t(1,h)}$ and derive a new set $\bar{\mathcal{V}}_{1,h}$.

\begin{align*}
   \bar{\mathcal{V}}_{1,h}=\{\bar{P}^{1}_{1,h},\bar{P}^{2}_{1,h},...,\bar{P}^{M}_{1,H})\}
\end{align*}
where for all $m\in \{1,2,\cdots, M\}$
\begin{align}
    \bar{P}^{(m)}_{1,h}(x)=\exists y W_{(s(1,h),t(1,h))}(x,y)\wedge {P}^{m}_{t(1,h)}(y)
\end{align}
We have $H$ heads and we can derive $H$ sets $\bar{\mathcal{V}}_{1,1},\bar{\mathcal{V}}_{1,2}, \cdots, \bar{\mathcal{V}}_{1,H}$.

Since there is a skip connection, the inputs to the FFN block in first layer includes not only the $\bar{\mathcal{V}}_{1,1},\bar{\mathcal{V}}_{1,2}, \cdots, \bar{\mathcal{V}}_{1,H}$, but also the ${\mathcal{V}}_{t}=\{{P}^{(1)}_{t},{P}^{(2)}_{t},\cdots,{P}^{(M)}_{t}\}$ for each $t \in \mathcal{N}\backslash \mathcal{N}_1$.

For each $t\in \mathcal{N} \backslash \mathcal{N}_1$, there are two cases.
\begin{itemize}
    \item \textbf{Case 1}: there are several indices $\{s(1,k_1),s(1,k_2),\cdots,s(1,k_a)\}\subset \mathcal{S}_1$ which represent the same index as $t$.
    \item \textbf{Case 2}: there are no indices which represent the same index as $t$.
\end{itemize}
We can derive the following set
\begin{align}
    {\mathcal{V}}_{1,t}=\{{P}^{1}_{1,t},{P}^{2}_{1,t},\cdots,{P}^{M}_{1,t}\}
\end{align}
, where for all $m\in \{1,2,\cdots, M\}$,
\begin{align}\label{new1V}
    {P}^{m}_{1,t}(x)=\left\{
\begin{array}{rcl}
{P}^{m}_{t}(x)\wedge \bar{P}^{m}_{1,k_{1}}(x) \wedge \cdots \wedge \bar{P}^{m}_{1,k_{t}}(x)      &      & \text{Case 1}\\
 {P}^{m}_{t}(x)       &      & \text{Case 2}
\end{array} \right. 
\end{align}

 the restricted variables ${y_n}$ where $n\in \mathcal{N}_1$, the target \ref{PNFtargetapp} will be transformed into the following formulation which preserves the logical equivalence.
\begin{align}\label{PNFtarget1}
        P(y_0) = \overset{} {\underset{n\in \mathcal{N} \backslash \mathcal{N}_1, n\neq 0}{\exists}} y_n \overset{M} {\underset{m=1}{\vee}}  \left[\left(\overset{} {\underset{t\in \mathcal{N} \backslash \mathcal{N}_1}{\wedge}} P^m_{1,t}(y_t)\right)\wedge \left(\overset{} {\underset{(t_p,t_c)\in {\mathcal{E}\backslash \mathcal{E}_1}}{\wedge}} W_{(t_p,t_c)}(y_{t_p},y_{t_c})\right)\wedge Q^m)\right]
\end{align}
, where $P^m_{1,t}$ is defined in \ref{new1V}.

\textbf{The Second Layer}

The second layer eliminate all the leaf nodes in $\mathcal{N}_2$. There are $H_2$ heads. 

For the $h$-th head in the first layer, the value matrix will represent the set of predicates $\mathcal{V}_{1,t(2,h)}=\{P^{1}_{1,t(2,h)},P^{2}_{1,t(2,h)},\cdots,P^{M}_{1,t(2,h)}\}$. Then the $h$-th head will learn the join operation between $W_{(s(2,h),t(2,h))}$ and $\mathcal{V}_{t(2,h)}$ and derive a new set $\bar{\mathcal{V}}_{2,h}$.

\begin{align*}
   \bar{\mathcal{V}}_{2,h}=\{\bar{P}^{1}_{2,h},\bar{P}^{2}_{2,h},...,\bar{P}^{M}_{2,h})\}
\end{align*}
where for all $m\in \{1,2,\cdots, M\}$
\begin{align}
    \bar{P}^{m}_{2,h}(x)=\exists y W_{(s(2,h),t(2,h))}(x,y)\wedge {P}^{m}_{1,t(2,h)}(y)
\end{align}
We have $H_2$ heads and we can derive $H_2$ sets $\bar{\mathcal{V}}_{2,1},\bar{\mathcal{V}}_{2,2}, \cdots, \bar{\mathcal{V}}_{2,H_2}$.

Since there is a skip connection, the inputs to the FFN block in first layer includes not only the $\bar{\mathcal{V}}_{2,1},\bar{\mathcal{V}}_{2,2}, \cdots, \bar{\mathcal{V}}_{2,H_2}$, but also the ${\mathcal{V}}_{1,t}=\{{P}^{(1)}_{1,t},{P}^{(2)}_{1,t},\cdots,{P}^{(M)}_{1,t}\}$ for each $t \in \mathcal{N}\backslash (\mathcal{N}_1\cup \mathcal{N}_2)$.

For each $t\in \mathcal{N} \backslash (\mathcal{N}_1\cup \mathcal{N}_2)$, there are two cases.
\begin{itemize}
    \item \textbf{Case 1}: there are several indices $\{s(2,k_1),s(2,k_2),\cdots,s(2,k_a)\}\subset \mathcal{S}_2$ which represent the same index as $t$.
    \item \textbf{Case 2}: there are no indices which represent the same index as $t$.
\end{itemize}
We can derive the following set
\begin{align}
    {\mathcal{V}}_{2,t}=\{{P}^{1}_{2,t},{P}^{2}_{2,t},\cdots,{P}^{M}_{2,t}\}
\end{align}
, where for all $m\in \{1,2,\cdots, M\}$,
\begin{align}\label{new1V2}
    {P}^{m}_{2,t}(x)=\left\{
\begin{array}{rcl}
{P}^{m}_{t}(x)\wedge \bar{P}^{m}_{2,k_{1}}(x) \wedge \cdots \wedge \bar{P}^{m}_{2,k_{t}}(x)      &      & \text{Case 1}\\
 {P}^{m}_{1,t}(x)       &      & \text{Case 2}
\end{array} \right. 
\end{align}
If we eliminate the restricted variables ${y_n}$ where $n\in \mathcal{N}_2$ in our target \ref{PNFtarget1}, the target \ref{PNFtarget1} will be transformed into the following formulation which preserves the logical equivalence.

\begin{align}\label{PNFtarget2}
        P(y_0) = \overset{} {\underset{n\in \mathcal{N} \backslash (\mathcal{N}_1\cup \mathcal{N}_2), n\neq 0}{\exists}} y_n \overset{M} {\underset{m=1}{\vee}}  \left[\left(\overset{} {\underset{t\in \mathcal{N} \backslash (\mathcal{N}_1\cup \mathcal{N}_2)}{\wedge}} P^m_{1,t}(y_t)\right)\wedge \left(\overset{} {\underset{(t_p,t_c)\in {\mathcal{E}\backslash (\mathcal{E}_1\cup \mathcal{E}_2)}}{\wedge}} W_{(t_p,t_c)}(y_{t_p},y_{t_c})\right)\wedge Q^m)\right]
\end{align}
, where $P^m_{2,t}$ is defined in \ref{new1V2}.

\textbf{The $l$-th Layer}

Iteratively, in the $l$-th layer, we eliminate all the leaf nodes in $\mathcal{N}_l$. There are $H_l$ heads.  

For the $h$-th head in the first layer, the value matrix will represent the set of predicates $\mathcal{V}_{l-1,t(l,h)}=\{P^{1}_{l-1,t(l,h)},P^{2}_{l-1,t(l,h)},\cdots,P^{M}_{1,t(l-1,h)}\}$. Then the $h$-th head will learn the join operation between $W_{(s(l,h),t(l,h))}$ and $\mathcal{V}_{l-1,t(l,h)}$ and derive a new set $\bar{\mathcal{V}}_{l,h}$.
\begin{align*}
   \bar{\mathcal{V}}_{l,h}=\{\bar{P}^{1}_{l,h},\bar{P}^{2}_{l,h},...,\bar{P}^{M}_{l,h})\}
\end{align*}
where for all $m\in \{1,2,\cdots, M\}$
\begin{align}
    P^{m}_{l,h}(x)=\exists y W_{(s(l,h),t(l,h))}(x,y)\wedge {P}^{m}_{t(l,h)}(y)
\end{align}
We have $H_l$ heads and we can derive $H_l$ sets $\bar{\mathcal{V}}_{l,1},\bar{\mathcal{V}}_{l,2}, \cdots, \bar{\mathcal{V}}_{l,H_l}$.

Since there is a skip connection, the inputs to the FFN block in first layer includes not only the $\bar{\mathcal{V}}_{l,1},\bar{\mathcal{V}}_{l,2}, \cdots, \bar{\mathcal{V}}_{l,H_l}$, but also the ${\mathcal{V}}_{l-1,t}=\{{P}^{(1)}_{l-1,t},{P}^{(2)}_{l-1,t},\cdots,{P}^{(M)}_{l-1,t}\}$ for each $t \in \mathcal{N}\backslash (\mathcal{N}_1\cup \mathcal{N}_2\cup \cdots \cup \mathcal{N}_{l-1} \cup \mathcal{N}_{l})$.

For each $t\in \mathcal{N} \backslash (\mathcal{N}_1\cup \mathcal{N}_2\cup \cdots \cup \mathcal{N}_{l-1}\cup \mathcal{N}_{l})$, 

there are two cases.
\begin{itemize}
    \item \textbf{Case 1}: there are several indices $\{s(l,k_1),s(l,k_2),\cdots,s(l,k_a)\}\subset \mathcal{S}_l$ which represent the same index as $t$.
    \item \textbf{Case 2}: there are no indices which represent the same index as $t$.
\end{itemize}
We can derive the following set
\begin{align}
    {\mathcal{V}}_{l,t}=\{{P}^{1}_{l,t},{P}^{2}_{l,t},\cdots,{P}^{M}_{l,t}\}
\end{align}
, where for all $m\in \{1,2,\cdots, M\}$,
\begin{align}\label{new1Vl}
    {P}^{m}_{l,t}(x)=\left\{
\begin{array}{rcl}
{P}^{m}_{l,t}(x)\wedge \bar{P}^{m}_{l,k_{1}}(x) \wedge \cdots \wedge \bar{P}^{m}_{l,k_{t}}(x)      &      & \text{Case 1}\\
 {P}^{m}_{l-1,t}(x)       &      & \text{Case 2}
\end{array} \right. 
\end{align}
If we eliminate the restricted variables ${y_n}$ where $n\in \mathcal{N}_l$, the target will be transformed into the following formulation which preserves the logical equivalence.
\begin{align}\label{PNFtargetl}
P(y_0) = \overset{} {\underset{n\in\mathcal{N} \backslash (\mathcal{N}_1\cup  \cdots \cup \mathcal{N}_{l-1}\cup \mathcal{N}_{l}), n\neq 0}{\exists}} y_n \overset{M} {\underset{m=1}{\vee}}  \Bigg[\left(\overset{} {\underset{t\in \mathcal{N} \backslash (\mathcal{N}_1\cup  \cdots \cup \mathcal{N}_{l-1}\cup \mathcal{N}_{l})}{\wedge}} P^m_{l-1,t}(y_t)\right)\wedge  \nonumber \\
 \left(\overset{} {\underset{(t_p,t_c)\in {\mathcal{E}\backslash (\mathcal{E}_1\cup \mathcal{E}_2\cup \cdots \cup \mathcal{E}_l)}}{\wedge}} W_{(t_p,t_c)}(y_{t_p},y_{t_c})\right)\wedge Q^m \Bigg],
\end{align} where $P^m_{l-1,t}(y_t)$ is defined in \ref{new1Vl}.

\textbf{The Last layer ($(L-1)$-th layer)}

After the self-attention block of last layer, we have eliminated all the variables except the root variable $y_0$.
The function of FFN in last layer is a little different than that in the general $l$-th layer. After deriving 
\begin{align}
    {\mathcal{V}}_{L-1,t}=\{{P}^{1}_{L-1,0},{P}^{2}_{K-1,0},\cdots,{P}^{M}_{L-1,0}\},
\end{align}
 our target \ref{PNFtargetl} can be transformed into 
\begin{align*}
    p(y_0)=({P}^{1}_{L-1,0}(y_{0})\wedge Q^1) \vee ({P}^{2}_{L-1,0}(y_{0})\wedge Q^2) \vee \cdots \vee ({P}^{M}_{L-1,0}(y_{0})\wedge Q^M)
\end{align*}

Then the FFN in the last layer can compute the result of our target function.

\textbf{General Case}

Our target is 
\begin{align}\label{genePNFtarget}
        P(y_0) = \exists y_1, \exists y_2, ..., \exists y_T\overset{M} {\underset{m=1}{\vee}}  (\overset{} {\underset{t\in {\mathcal{N}^m}}{\wedge}} P^m_{t}(y_t)\wedge \overset{} {\underset{(t_p,t_c)\in {\mathcal{E}^m}}{\wedge}} W^i_{(t_p,t_c)}(y_{t_p},y_{t_c})\wedge Q_m)
\end{align}

For each $\mathcal{N}^m$, we can derive the following partition in the same way as the simple case.
\begin{align*}
    \mathcal{N}^m=\mathcal{N}^m_1 \cup \mathcal{N}^m_2 \cup \cdots \cup \mathcal{N}^m_{L^m} 
\end{align*}
, where $\mathcal{N}^m_j =\{t^m{(j,1)},t^m{(j,2)},\cdots, t^m{(j,h_j^m)}\}$ is the set of leaf nodes.
Then we can derive the following group for $j\in \{1,2,\cdots, L_{max}\}$. \textbf{Here we abuse the definition of set since there might be duplicate elements in $\mathcal{N}_j$.}
\begin{align*}
    \mathcal{N}_j=\mathcal{N}^1_j \cup \mathcal{N}^2_j \cup \cdots \cup \mathcal{N}^M_j
\end{align*}
, where $\mathcal{N}^m_j=\emptyset$ for $L^m<j\leq L_{max}$.

Similarly, we can also derive the $\mathcal{S}^m_j$ for each $m$, which is the set of parent nodes which is connected to the leaf nodes in $\mathcal{N}^m_j$.
\begin{align*}
    \mathcal{S}^m_j =\{s^m{(j,1)},s^m{(j,2)},\cdots, s^m{(j,h_j^m)}\}
\end{align*}
Then we can derive the following group for $j\in \{1,2,\cdots, L_{max}\}$. \textbf{Here we abuse the definition of set since there might be duplicate elements in $\mathcal{S}_j$.}
\begin{align*}
    \mathcal{S}_j=\mathcal{S}^1_j \cup \mathcal{S}^2_j \cup \cdots \cup \mathcal{NS}^M_j
\end{align*}
, where $\mathcal{S}^m_j=\emptyset$ for $L^m<j\leq L_{max}$.

As for each edge set $\mathcal{E}^m$, we can derive the following partition in the same way as the simple case.
\begin{align*}
    \mathcal{E}^m=\mathcal{E}^m_1 \cup \mathcal{E}^m_2 \cup \cdots \cup \mathcal{E}^m_{L^m} 
\end{align*}
And 
\begin{align*}
    \mathcal{E}^m_j =\{(s^m{(j,1)},t^m{(j,1)}),(s^m{(j,2)},t^m{(j,2)}),\cdots, (s^m{(j,h_1^m)},t^m{(j,h_1^m)})\}
\end{align*}
 is the set of leaf nodes.
Then we can derive the following group for $j\in \{1,2,\cdots ,L_{max}\}$.
\begin{align*}
    \mathcal{E}_j=\mathcal{E}^1_j \cup \mathcal{E}^2_j \cup \cdots \cup \mathcal{E}^M_j
\end{align*}
, where $\mathcal{E}^m_j=\emptyset$ for $L^m<j\leq L_{max}$.

\textbf{The First Layer}
Similar to the proof for the simple cases, the first layer eliminate all the leaf nodes in $\mathcal{N}_1$. There are $h^1_1+h^2_1+\cdots+h^M_1$ heads. 

For the $k$-th head in the first layer, where $h^1_1+...+h^{m_k-1}_1<k\leq h^1_1+...+h^{m_k}_1$, it will eliminate the $k$-th node in $\mathcal{N}_1$. We denote that node as $z(1,k)$ and its parent node in $\mathcal{S}_1$ as $r(1,k)$. The value matrix will include the predicate $P^{m_k}_{z(1,k)}$. 
Then the $k$-th head will learn the join operation between $W^{m_k}_{(r(1,k),z(1,k))}$ and $P^{{m_k}}_{z(1,k)}$ and derive a new predicate $\bar{P}^{m_k}_{1,k}$.
\begin{align*}
    \bar{P}^{m_k}_{1,k}(x)=\exists y W^{m_k}_{(r(1,k),z(1,k))}(x,y) \wedge P^{{m_k}}_{z(1,k)}(y)
\end{align*}
We have $h^1_1+h^2_1+\cdots+h^M_1$ heads so we can have $h^1_1+h^2_1+\cdots+h^M_1$ predicates $\{\bar{P}^{m_k}_{1,k}\}^{h^1_1+h^2_1+\cdots+h^M_1}_{k=1}$ mentioned above.

Since there is a skip connection, the inputs to the FFN block in first layer includes not only the $\{\bar{P}^{m_k}_{1,k}\}^{h^1_1+h^2_1+\cdots+h^M_1}_{k=1}$, but also the $\{P^m_{t}\}$ for each $t \in \mathcal{N}^m \backslash \mathcal{N}^m_1$ and $1 \leq m\leq M$.

For each $1 \leq m\leq M$ and $t\in \mathcal{N}^m \backslash \mathcal{N}^m_1$, if there are several indices $\{r(1,k_1),r(1,k_2),\cdots,r(1,k_a)\}\subset \mathcal{S}^m_1$ which represent the same index as $t$ and $m_{k_1}=m_{k_2}=\cdots=m$, we can derive the following predicate
\begin{align}\label{genenew1V}
    {P}^{m}_{1,t}(x)={P}^{m}_{t}(x)\wedge P^{m_{k_{1}}}_{1,k_{1}}(x) \wedge \cdots \wedge P^{m_{k_{t}}}_{1,k_{t}}(x).
\end{align}
For each $m$ and $t\in \mathcal{N}^m \backslash \mathcal{N}^m_1$, if there are no indices in $\mathcal{S}^m_1$ which represents the same index as $t$, we can derive the following predicate
\begin{align} \label{genenew2V}
  {P}^{m}_{1,t}(x)={P}^{m}_{t}(x)
\end{align}

If we eliminate the restricted variables ${y_n}$ where $n\in \mathcal{N}^m_1$, the target \ref{genePNFtarget} will be transformed into the following formulation which preserves the logical equivalence.
\begin{align}\label{genePNFtarget1}
        P(y_0) =  \overset{M} {\underset{m=1}{\vee}}  ({\underset{n\in \mathcal{N^m} \backslash \mathcal{N^m}_1, n\neq 0}{\exists}} y_n {\underset{t\in \mathcal{N^m} \backslash \mathcal{N^m}_1}{\wedge}} P^m_{1,t}(y_t)\wedge \overset{} {\underset{(t_p,t_c)\in {\mathcal{E^m}\backslash \mathcal{E^m}_1}}{\wedge}} W^m_{(t_p,t_c)}(y_{t_p},y_{t_c})\wedge Q)
\end{align}
, where $P^m_{1,t}$ is defined in \ref{genenew1V} and \ref{genenew2V}.

\textbf{The Second Layer}
Similar to the proof for the simple cases, the second layer eliminate all the leaf nodes in $\mathcal{N}_2$. There are $h^1_2+h^2_2+\cdots+h^M_2$ heads. 

For the $k$-th head in the second layer, where $h^1_2+...+h^{m_k-1}_2<k\leq h^1_2+...+h^{m_k}_2$, it will eliminate the $k$-th node in $\mathcal{N}_2$. We denote that node as $z(2,k)$ and its parent node in $\mathcal{S}_2$ as $r(2,k)$. The value matrix will include the predicate $P^{m_k}_{z(2,k)}$. 
Then the $k$-th head will learn the join operation between $W^{m_k}_{(r(2,k),z(2,k))}$ and $P^{{m_k}}_{z(2,k)}$ and derive a new predicate $\bar{P}^{m_k}_{2,k}$.
\begin{align*}
    \bar{P}^{m_k}_{2,k}(x)=\exists y W^{m_k}_{(r(2,k),z(2,k))}(x,y) \wedge P^{{m_k}}_{z(2,k)}(y)
\end{align*}

We have $h^1_2+h^2_2+\cdots+h^M_2$ heads so we can have $h^1_2+h^2_2+\cdots+h^M_2$ predicates $\{\bar{P}^{m_k}_{2,k}\}^{h^1_2+h^2_2+\cdots+h^M_2}_{k=1}$ mentioned above.

Since there is a skip connection, the inputs to the FFN block in first layer includes not only the $\{\bar{P}^{m_k}_{2,k}\}^{h^1_2+h^2_2+\cdots+h^M_2}_{k=1}$, but also the $\{P^m_{1,t}\}$ for each $t \in \mathcal{N}^m \backslash (\mathcal{N}^m_1 \cup \mathcal{N}^m_2)$ and $1 \leq m\leq M$.

For each $1 \leq m\leq M$ and $t\in \mathcal{N}^m \backslash (\mathcal{N}^m_1 \cup \mathcal{N}^m_2)$, if there are several indices $\{r(2,k_1),r(2,k_2),\cdots,r(2,k_a)\}\subset \mathcal{S}^m_2$ which represent the same index as $t$ and $m_{k_1}=m_{k_2}=\cdots=m_{k_a}=m$, we can derive the following predicate
\begin{align}\label{genenew1Vl}
    {P}^{m}_{2,t}(x)={P}^{m}_{1,t}(x)\wedge \bar{P}^{m_{k_{1}}}_{2,k_{1}}(x) \wedge \cdots \wedge \bar{P}^{m_{k_{t}}}_{2,k_{t}}(x).
\end{align}
For each $m$ and $t\in \mathcal{N}^m \backslash \mathcal{N}^m_1$, if there are no indices in $\mathcal{S}^m_1$ which represents the same index as $t$, we can derive the following predicate
\begin{align} \label{genenew2Vl}
  {P}^{m}_{2,t}(x)=\bar{P}^{m}_{1,t}(x)
\end{align}

If we eliminate the restricted variables ${y_n}$ where $n \in \mathcal{N}^m_2$, the target \ref{genePNFtarget1} will be transformed into the following formulation which preserves the logical equivalence.
\begin{align}\label{genePNFtarget2}
        P(y_0) =  \overset{M} {\underset{m=1}{\vee}}  ({\underset{n\in \mathcal{N}^m \backslash (\mathcal{N}^m_1 \cup \mathcal{N}^m_2), n\neq 0}{\exists}} y_n {\underset{t\in \mathcal{N}^m \backslash \mathcal{N}^m_1}{\wedge}} P^m_{2,t}(y_t)\nonumber \\
        \wedge \overset{} {\underset{(t_p,t_c)\in {\mathcal{E}^m\backslash (\mathcal{E}^m_1 \cup \mathcal{E}^m_2)}}{\wedge}} W^m_{(t_p,t_c)}(y_{t_p},y_{t_c})\wedge Q)
\end{align}
, where $P^m_{1,t}$ is defined in \ref{genenew1Vl} and \ref{genenew2Vl}.

\textbf{The $l$-th Layer}
Similar to the proof for the simple cases, the $l$-th layer eliminate all the leaf nodes in $\mathcal{N}_l$. There are $h^1_l+h^2_l+\cdots+h^M_l$ heads. 

For the $k$-th head in the $l$-th layer, where $h^1_l+...+h^{m_k-1}_l<k\leq h^1_l+...+h^{m_k}_l$, it will eliminate the $k$-th node in $\mathcal{N}_l$. We denote that node as $z(l,k)$ and its parent node in $\mathcal{S}_l$ as $r(l,k)$. The value matrix will include the predicate $P^{m_k}_{z(l,k)}$. 
Then the $k$-th head will learn the join operation between $W^{m_k}_{(r(l,k),z(l,k))}$ and $P^{{m_k}}_{z(l,k)}$ and derive a new predicate $\bar{P}^{m_k}_{l,k}$.
\begin{align*}
    \bar{P}^{m_k}_{l,k}(x)=\exists y W^{m_k}_{(r(l,k),z(l,k))}(x,y) \wedge P^{{m_k}}_{z(l,k)}(y)
\end{align*}

We have $h^1_l+h^2_l+\cdots+h^M_l$ heads so we can have $h^1_l+h^2_l+\cdots+h^M_l$ predicates $\{\bar{P}^{m_k}_{l,k}\}^{h^1_l+h^2_l+\cdots+h^M_l}_{k=1}$ mentioned above.

Since there is a skip connection, the inputs to the FFN block in first layer includes not only the $\{\bar{P}^{m_k}_{l,k}\}^{h^1_l+h^2_l+\cdots+h^M_l}_{k=1}$, but also the $\{P^m_{l-1,t}\}$ for each $t \in \mathcal{N}^m \backslash (\mathcal{N}^m_1 \cup \mathcal{N}^m_2\cup \cdots \cup \mathcal{N}^m_{l})$ and $1 \leq m\leq M$.

For each $1 \leq m\leq M$ and $t\in \mathcal{N}^m \backslash (\mathcal{N}^m_1 \cup \mathcal{N}^m_2\cup \cdots \cup \mathcal{N}^m_{l})$, if there are several indices $\{r(l,k_1),r(l,k_2),\cdots,r(l,k_a)\}\subset \mathcal{S}^m_l$ which represent the same index as $t$ and $m_{k_1}=m_{k_2}=\cdots=m_{k_a}=m$, we can derive the following predicate
\begin{align}\label{genenew1V2}
    {P}^{m}_{l,t}(x)={P}^{m}_{l-1,t}(x)\wedge \bar{P}^{m_{k_{1}}}_{l,k_{1}}(x) \wedge \cdots \wedge \bar{P}^{m_{k_{t}}}_{l,k_{t}}(x).
\end{align}
For each $m$ and $t\in \mathcal{N}^m \backslash \mathcal{N}^m_1$, if there are no indices in $\mathcal{S}^m_1$ which represents the same index as $t$, we can derive the following predicate
\begin{align} \label{genenew2V2}
  {P}^{m}_{l,t}(x)=\bar{P}^{m}_{l-1,t}(x)
\end{align}

If we eliminate the restricted variables ${y_n}$ where $n \in \mathcal{N}^m_l$, the target \ref{genePNFtarget1} will be transformed into the following formulation which preserves the logical equivalence.
\begin{align}\label{genePNFtarget2}
        P(y_0) =  \overset{M} {\underset{m=1}{\vee}}  ({\underset{n\in \mathcal{N}^m \backslash (\mathcal{N}^m_1 \cup \mathcal{N}^m_2\cup \cdots \cup \mathcal{N}^m_{l}), n\neq 0}{\exists}} y_n {\underset{t\in \mathcal{N}^m \backslash (\mathcal{N}^m_1 \cup \mathcal{N}^m_2\cup \cdots \cup \mathcal{N}^m_{l})}{\wedge}} P^m_{l,t}(y_t)\nonumber \\
        \wedge \overset{} {\underset{(t_p,t_c)\in {\mathcal{E}^m\backslash (\mathcal{E}^m_1 \cup \mathcal{E}^m_2\cup \cdots \cup \mathcal{E}^m_{l})}}{\wedge}} W^m_{(t_p,t_c)}(y_{t_p},y_{t_c})\wedge Q)
\end{align}
, where $P^m_{1,t}$ is defined in \ref{genenew1Vl} and \ref{genenew2Vl}.

\textbf{The Last layer ($(L-1)$-th layer)}

After the self-attention block of last layer, we have eliminated all the variables except the root variable $y_0$.
The function of FFN in last layer is a little different than that in the general $l$-th layer. After deriving 
\begin{align}
    {P}^{m}_{L-1,0}(y_0)
\end{align} for $1\leq m \leq M$
, our target \ref{PNFtargetl} can be transformed into 

\begin{align*}
    p(y_0)={P}^{1}_{L-1,0}(y_{0}) \vee {P}^{2}_{L-1,0}(y_{0}) \vee \cdots \vee {P}^{M}_{L-1,0}(y_{0})
\end{align*}

Then the FFN in the last layer can compute the result of our target.

\end{document}